%% file: main.tex
\documentclass[a4paper, 10pt, conference, final]{ieeeconf}      

\IEEEoverridecommandlockouts                              
\usepackage{siunitx}
\usepackage{subcaption}
\usepackage{wrapfig}
\usepackage{amsmath}
\usepackage{mathtools}
\usepackage{tikz}
\usetikzlibrary{calc}
\usepackage{mwe}
\usetikzlibrary{arrows,decorations,backgrounds,calc}
\usepackage{relsize}
\usepackage{pgfplots}
\usepgfplotslibrary{groupplots}
\pgfplotsset{compat=1.14}
\usepackage{booktabs}
\usepackage[noadjust]{cite}

\usepackage{hyperref}
\usepackage[capitalise]{cleveref}

\tikzset {> = stealth, 
fontscale/.style = {font=\relsize{#1}},
  edge rectangle/.style={
    to path={ rectangle (\tikztotarget)}
  }
}

\usepackage{todonotes}
\usepackage{soul}
\definecolor{smoothgreen}{rgb}{0.7,1,0.7}
\sethlcolor{smoothgreen}

\title{\LARGE \bf Dynamic Environment Prediction in Urban Scenes \\ using Recurrent Representation Learning}

\author{Masha Itkina,$^{1}$ Katherine Driggs-Campbell,$^{2}$ and Mykel J. Kochenderfer$^{1}$
\thanks{This work was supported by the Ford-Stanford Alliance.}
\thanks{$^{1}$M. Itkina and M. J. Kochenderfer are with the Aeronautics and Astronautics Department, Stanford University, 496 Lomita Mall, Stanford, CA 94305, USA. Email: {\tt\small \{mitkina,mykel\}@stanford.edu}}%
\thanks{$^{2}$K. Driggs-Campbell is with the Electrical and Computer Engineering Department, University of Illinois at Urbana-Champaign, 306 N. Wright St. MC 702 Urbana, IL 61801, USA. Email: {\tt\small krdc@illinois.edu}}%
}

\begin{document}

\maketitle

\thispagestyle{empty}
\pagestyle{empty}

\begin{abstract}

A key challenge for autonomous driving is safe trajectory planning in cluttered, urban environments with dynamic obstacles, such as pedestrians, bicyclists, and other vehicles. 
A reliable prediction of the future environment, including the behavior of dynamic agents, would allow planning algorithms to proactively generate a trajectory in response to a rapidly changing environment. 
We present a novel framework that predicts the future occupancy state of the local environment surrounding an autonomous agent by learning a motion model from occupancy grid data using a neural network.
We take advantage of the temporal structure of the grid data by utilizing a convolutional long-short term memory network in the form of the PredNet architecture.
This method is validated on the KITTI dataset and demonstrates higher accuracy and better predictive power than baseline methods.
\end{abstract}

\input{Sections/01-Intro}
\input{Sections/03-Methods}
\input{Sections/04-Experiments}
\input{Sections/05-Results}
\input{Sections/06-Conclusion}
\input{Sections/07-Acknowledgements}

\bibliographystyle{IEEEtran}
\bibliography{BibFile}

\end{document}

%% file: Sections/01-Intro.tex
\section{INTRODUCTION} \label{sec:intro}
One of the key challenges in autonomous driving is maneuvering in cluttered environments in the presence of other road users, such as moving vehicles, pedestrians, and bicyclists. Conventionally, planning and control algorithms, such as model predictive control, plan a trajectory over some time horizon using the current environment state~\cite{mpc_orig}.
Information regarding the future environment, specifically the behavior of other dynamic agents, would facilitate planning a trajectory proactively to a rapidly changing environment~\cite{mpc_good_models}. For instance, in the context of autonomous vehicles, accurate environment prediction would provide for a smoother user experience, with fewer sharp maneuvers, while ensuring passenger safety from possible moving hazards on the road~\cite{mpc_gerdes}.

To make intelligent predictions, the autonomous vehicle must generate a representation of the state of the environment from sensor data, commonly in the form of a map.
Mapping techniques can be categorized into
discrete and continuous representations.
Continuous representations generally take the form of an object or landmark list within the environment~\cite{siegwart_mobile_robots}. 
Behavior models can then be used to predict the future state of detected dynamic objects (e.g., using the intelligent driver model for autonomous vehicle applications~\cite{idm}). 
This approach implies extensive knowledge of the environment and of the agents' likely behaviors. 
Furthermore, these behavior models are limited in their generalization capabilities by considering only certain contexts and scenarios~\cite{siegwart_mobile_robots,meyer_dynamic_occupancy,dietmayer}.

The alternative representation is a discretized occupancy grid, which is a common method for mobile robotics applications~\cite{occupancy_grids}. 
Occupancy grids discretize the world into independent cells. Each cell contains the belief that the discrete space is occupied. 
In contrast to continuous approaches, occupancy grids can generate a probabilistic map without having knowledge of the entire environment or making assumptions regarding the agent behavior by directly incorporating raw sensor measurements~\cite{hoermann_object_extraction}. 
Furthermore, occupancy grids facilitate the use of common perception techniques such as object detection~\cite{hoermann_detection} and tracking~\cite{hoermann_deep_tracking}. 
Bayesian methods are often used to incorporate sensor measurements into occupancy grids~\cite{occupancy_grids}. Alternatively, Dempster-Shafer Theory (DST) can be used to provide evidential updates that combine elements of evidence in support of or against a set of hypotheses.
DST is a decision-making strategy that differentiates lack of information from conflicting or uncertain information~\cite{dst}. 
For example, a Bayesian grid may output a $0.5$ occupancy probability for a cell that did not receive a sensor measurement (i.e., an occluded region) and for a cell that was occupied but is now free (i.e., a moving object). 
DST effectively separates these two cases to distinguish moving obstacles from occluded space, making the approach well suited for dynamic settings~\cite{kurdej}. DST's modularity makes it particularly adept at combining multiple sensor measurements in sensor-fusion tasks~\cite{sensor_fusion_2002,sensor_fusion_mobile_robots_1998,sensor_fusion_IV,sensor_fusion_airborne}.

Even though these mapping paradigms provide a spatial representation for the environment, they do not directly address the dynamics of the agents within it. We aim to design a predictive model to facilitate downstream planning for an autonomous vehicle in a cluttered, urban setting. We use the evidential occupancy grid representation in conjunction with state-of-the-art computer vision techniques to formulate an estimate of the state of the environment at a future time.

\begin{figure*}[t!]
    \centering
    \scalebox{0.8}{\input{figs/process.tex}}
    \caption{\small Pipeline for the proposed framework.}
    \label{fig:methodology}
     \vspace{-10pt}
\end{figure*}
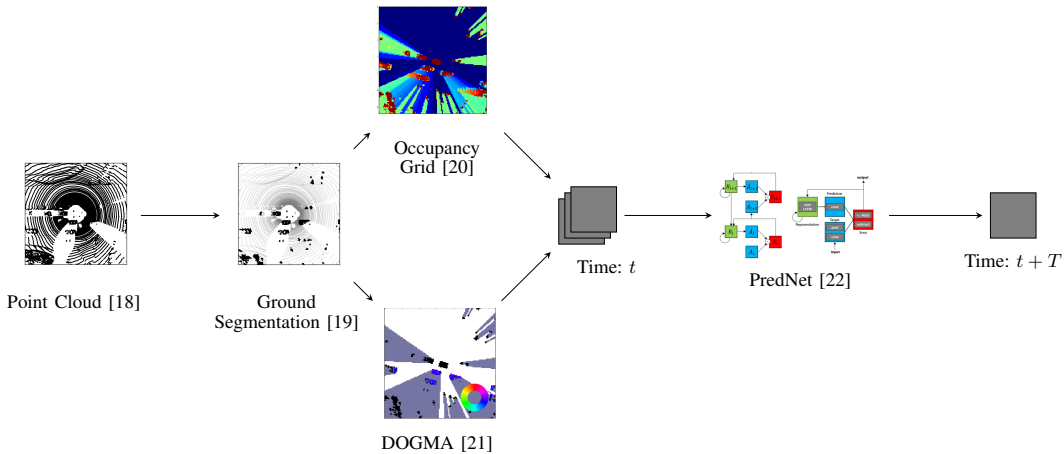

The perception community has developed several approaches for predicting dynamic environments.
For instance, a dynamic occupancy grid map (DOGMa) stores the dynamic state of each cell (e.g., velocity) 
in addition to the occupancy belief~\cite{nuss_particles,dietmayer}. 
The dynamic state can facilitate environment prediction given some transition model, which can propagate the occupancy to estimate the future environment state~\cite{rummelhard2015conditional}.

There are several approaches for generating DST-based DOGMas. Kurdej~et~al. present an algorithm to discern moving and static obstacles on the road by combining LiDAR sensor information with semantic, street-level GPS data~\cite{kurdej},~\cite{kurdejGEO}. 
Nuss et al. present a DST approximation for Bayesian Monte Carlo particle filtering that generates velocity-level information for each grid cell~\cite{nuss_particles,nuss_fusion}.

Recently, computer vision has been revolutionized by deep learning~\cite{deep_learning}.
For example, deep learning architectures have been developed to address the problems of semantic segmentation and video frame prediction. 
Semantic segmentation involves fully convolutional neural network architectures that are used to make pixel-wise classifications of images~\cite{fcn, deconv}.
Convolutional long short-term memory (ConvLSTM) networks are now a standard for temporal or sequential data, such as in video frame prediction~\cite{srivastava,prednet,goodfellow,spatio_temporal}. 

One advantage of a DOGMa representation is that it holds many parallels with RGB images. 
The dynamic state information is held in different channels akin to the RGB channels in images. The spatial representation of cells in an occupancy grid is akin to the association of neighboring pixels into objects in images.
Given these similarities, there are several works that learn an occupancy grid data representation via image-based deep learning techniques. Piewak et al. employ a fully convolutional neural network with DOGMa input to classify static and moving cells~\cite{fcn_baseline}. 
Hoermann et al. use DOGMas to predict and classify static and moving cells via the DeconvNet architecture, equating the problem of environment prediction to image segmentation~\cite{dietmayer}. 

Building upon this literature, we aim to predict future occupancy grids in cluttered, urban environments by using concepts from image-based deep-learning. We pose the environment state prediction problem as a video frame prediction task. 
We use a ConvLSTM architecture to take advantage of the temporal nature of occupancy grid data and learn both the temporal and spatial patterns in the grids to predict future environment state. By effectively merging dynamic map representations with deep learning architectures, we demonstrate how to learn a predictive environment model in cluttered settings. The proposed pipeline is applied to an autonomous, urban driving scenario and demonstrates improved environment prediction over baseline methods.

Several recent papers consider an LSTM network on occupancy grid data.
Kim et al. use a custom LSTM approach with probabilistic dynamic occupancy grids to predict highway vehicle trajectories with promising results~\cite{kim}.
LiDAR-FlowNet is another approach that estimates flow in the occupancy grid space to predict future maps, demonstrated on an indoor mobile robot~\cite{song20192d}.
We consider a more complex environment with multiple behavior modalities (e.g., vehicles, bicyclists, pedestrians) and structural components (e.g., buildings, intersections) than these scenarios.

Prior research has focused on object tracking from grid representations with the use of ConvLSTMs. 
Engel et al. exploit a ConvLSTM architecture for object tracking using DOGMas~\cite{hoermann_deep_tracking}. 
Luo et al. present a convolutional network architecture that jointly reasons about detection, tracking, and motion forecasting of dynamic agents~\cite{luo2018fast}. 
While they effectively detect and predict the motion of these agents, the work is not concerned with the evolution of the full environment as considered here. 
Dequaire et al. and Ondruska et al. address occlusion inference on a moving vehicle platform~\cite{deep_tracking,deep_tracking_in_wild}. In contrast, we focus on the task of ego-centric environment prediction using a ConvLSTM network and show the potential to streamline the pipeline by performing end-to-end learning on ordinary occupancy grids.

There are two concurrent works that address environment prediction from occupancy grids with ConvLSTMs. Schreiber et al. present a two-channel neural network architecture, separating the static and dynamic cells in DOGMas, and test on a stationary vehicle platform~\cite{schreiber}. Mohajerin and Rohani present a difference-learning architecture with computer vision-based motion features on a driving vehicle platform~\cite{huawei}. Despite a moving platform, the latter work makes a static ego vehicle assumption for the purposes of prediction. Both works consider an encoder-decoder architecture where the LSTM learns on the encoded features.

In contrast, we propose a model-free approach to environment prediction using the PredNet architecture, which builds an internal environment representation with time and outperforms standard encoder-decoder models~\cite{prednet}. We model the relative interactions between the ego vehicle and the environment; thus, we consider a local environment representation centered at a moving ego vehicle.

\begin{figure}[t!] 
	\centering
    \scalebox{0.65}{\input{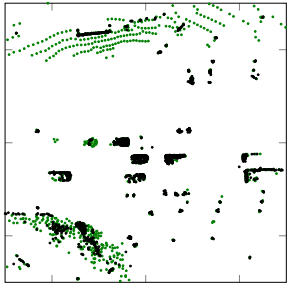}}
\caption{\small Processed point cloud from the KITTI dataset after filtering out the ground points using MRF and RANCSAC algorithms. Ground points missed by RANSAC but filtered out by MRF are highlighted in green.
} \label{fig:MRF}
\vspace{-10pt}
\end{figure}

Our contributions in this work are three-fold:

\begin{enumerate}
\item By framing the problem of environment prediction in an urban, cluttered setting as a video frame prediction task, we validate the capacity of the network to predict the environment multiple time steps into the future.
\item We show that the ConvLSTM learns an internal dynamic representation of the environment allowing for prediction from occupancy grid data without additional dynamic state information. 
\item We compare the benefits of a Dempster-Shafer environment representation to a probabilistic alternative.
\end{enumerate}

The paper is organized as follows. \Cref{sec:approach} outlines the details of our proposed approach.
\Cref{sec:experiments} describes the experimental procedure to validate our methodology. \Cref{sec:results} discusses experimental results. \Cref{sec:conclusions} draws conclusions and suggests directions for future research.

%% file: figs/process.tex
\begin{tikzpicture}[node distance = 3.5cm, auto]
\node(pointcloud) [label={[align=center]below: \small Point Cloud \cite{kitti}}] {\includegraphics[width=0.11\textwidth,trim={3cm 0cm 3cm 0cm},clip]{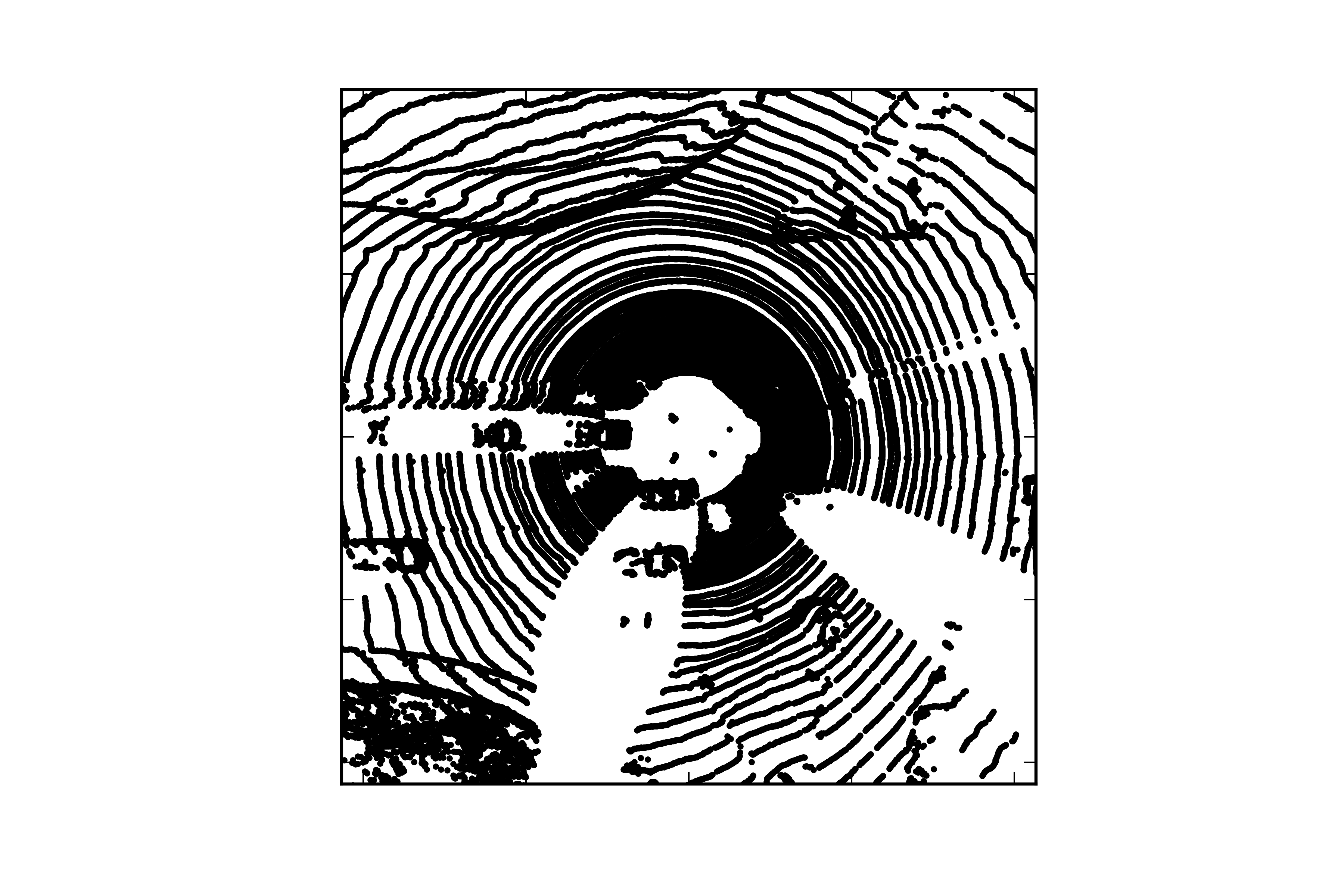}};

\node(groundseg) [label={[align=center]below:  \small Ground \\[-0.1cm] \small Segmentation~\cite{postica}}] [right of=pointcloud] {\includegraphics[width=0.11\textwidth,trim={3cm 0cm 3cm 0cm},clip]{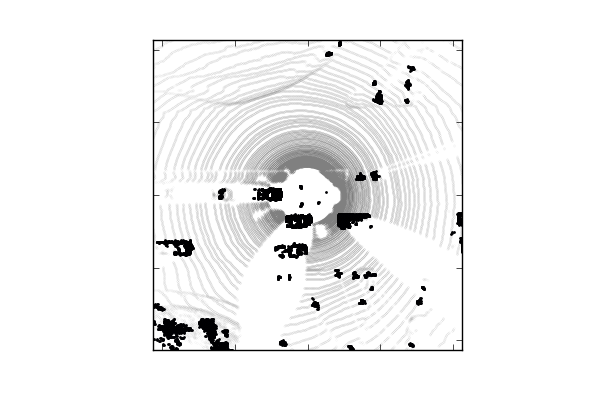}};


\node(static) [label={[align=center]below:  \small Occupancy \\[-0.1cm] \small Grid \cite{nuss_fusion}}] [above right of=groundseg] {\includegraphics[width=0.11\textwidth,trim={2.15cm 0cm 0cm 2cm},clip]{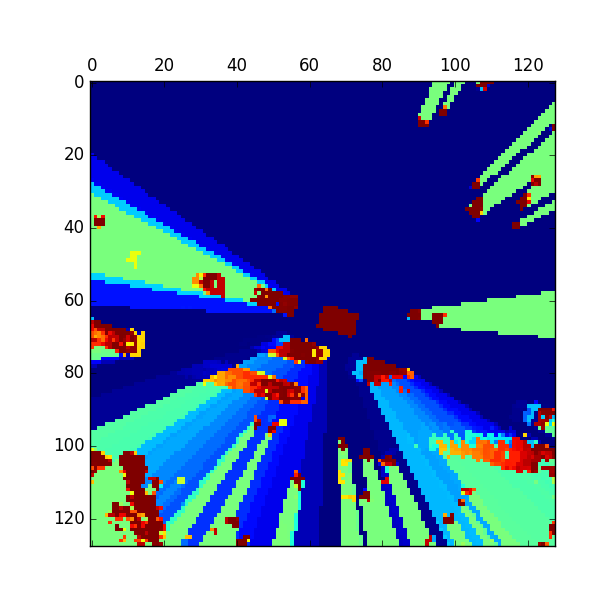}};

\node(dogma) [label={[align=center]below: \small DOGMA \cite{nuss_particles}}] [below right of=groundseg] {\includegraphics[width=.12\textwidth,trim={3cm 1.2cm 3cm 1.5cm},clip]{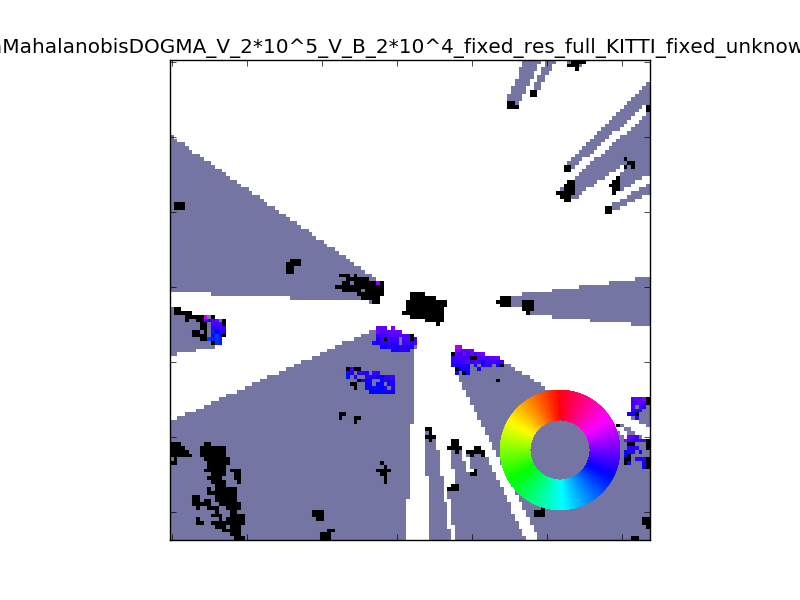}};

\node(prevt) [label={[align=center, xshift=2ex]below: \small Time: $t$}] [above right of=dogma] {
  \begin{tikzpicture}[anchor=center]
  \draw[every edge/.append style={edge rectangle,fill=gray}] (0,0)
  edge +(0.75,0.75) ++(0.1,0.1)
  edge +(0.75,0.75) ++(0.1,0.1)
  edge +(0.75,0.75) ++(0.1,0.1);
  \end{tikzpicture}
  };


\node(neuralnet) [label={[align=center]below: \small PredNet \cite{prednet}}] [right of=prevt] {\includegraphics[width=.15\textwidth,trim={0.75cm 0.5cm 0.75cm 0.75cm},clip]{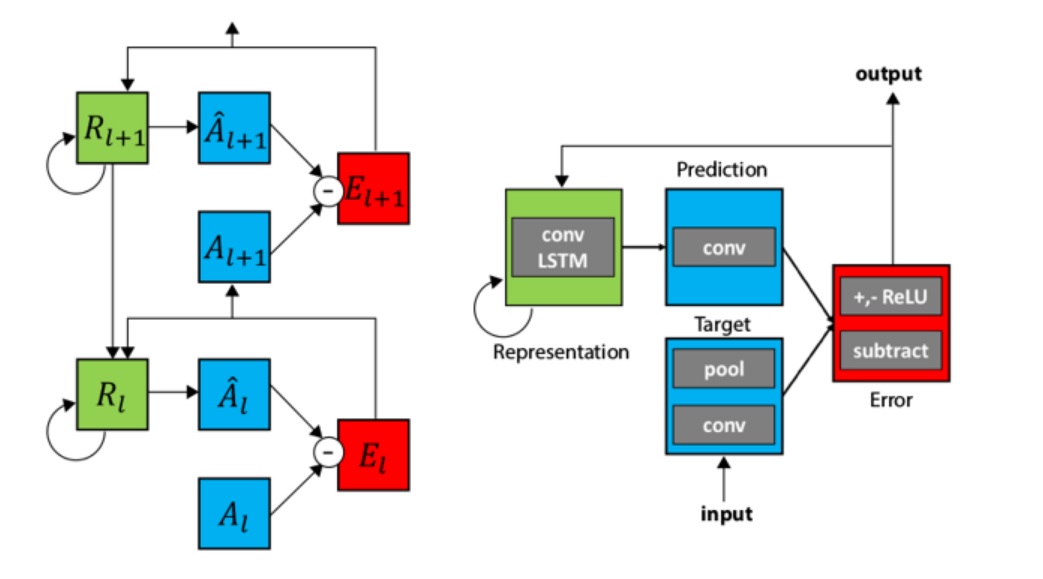}};

\node(nextt) [label={[align=center]below: \small Time: $t+T$ }] [right of=neuralnet] {
  \begin{tikzpicture}[anchor=center]
  \draw[every edge/.append style={edge rectangle,fill=gray}] (0,0)
  edge +(0.75,0.75) ++(0.1,0.1);
  \end{tikzpicture}
  };

\draw [->] (pointcloud) -- (groundseg);
\draw [->] (groundseg) -- (static);
\draw [->] (groundseg) -- (dogma);
\draw [->] (static) -- (prevt);
\draw [->] (dogma) -- (prevt);
\draw [->] (prevt) -- (neuralnet);
\draw [->] (neuralnet) -- (nextt);

\end{tikzpicture}

%% file: Sections/03-Methods.tex
\section{APPROACH} \label{sec:approach}

We use a ConvLSTM architecture intended for video frame prediction to instead predict the local environment surrounding an autonomous agent across future time steps.
For this purpose, we adapt the PredNet architecture designed for video frame prediction in autonomous driving scenes~\cite{prednet}.
The ConvLSTM is expected to learn an internal representation of the dynamics within the local environment from occupancy grid data. The grid input are generated from LiDAR measurement recordings in the KITTI dataset taken across a variety of urban scenes~\cite{kitti}. 
We investigate whether the learned representation is sufficient for prediction without additional cell-wise velocity estimation. For this purpose, we train PredNet on two different input types: static occupancy grids and DOGMas. We also investigate the effect of DST-based versus probabilistic grids as input into the neural network architecture.

The following sections describe the proposed pipeline and detail the input data pre-processing as well as the PredNet architecture. The process is outlined in \cref{fig:methodology}. 

\subsection{Ground Segmentation}
Prior to generating an occupancy grid, the ground must be segmented and removed from the LiDAR point cloud. 
Traditional ground segmentation methods, such as RANSAC, estimate the ground as a plane across a LiDAR scan.
Postica et al. present a Markov Random Field (MRF) algorithm that exploits local spatial relationships, avoiding the assumption of a planar ground~\cite{postica}.
Empirically, the MRF approach exhibits significantly improved ground segmentation compared to RANSAC.
An example segmentation on a LiDAR point cloud is visualized in \cref{fig:MRF}. The MRF successfully filters the radial ground points missed by RANSAC. 

\subsection{Dynamic Occupancy Grid Maps (DOGMas)}
A DOGMa is an evidential grid containing both occupancy and dynamic state information (e.g., velocity). We generate DOGMas via the procedure outlined by Nuss et al.~\cite{nuss_particles, nuss_fusion}. There are two parallel processes that occur: occupancy grid updates and cell-wise velocity estimates. We briefly describe these algorithms in the following sections.

\subsubsection{Occupancy Grids} \label{sec:dst_occ}
We consider DST-based occupancy grids computed from LiDAR measurements as detailed by Nuss et al.~\cite{nuss_fusion}. DST deals with a set of exhaustive hypotheses formed from a \textit{frame of discernment} and the associated belief masses. In the case of an occupancy grid, our frame of discernment is: $\Omega = \left\{F,O\right\}$, where $F$ is free space and $O$ is occupied space. Thus, the set of hypotheses is: $\left\{\emptyset, \left\{F\right\}, \left\{O\right\}, \left\{F,O\right\}\right\}$. The null set $\emptyset$ is impossible in this context as a cell physically cannot be neither occupied nor unoccupied. Thus, our exhaustive set of hypotheses is: $\left\{\left\{F\right\}, \left\{O\right\}, \left\{F,O\right\}\right\}$. The sum of the belief masses over the possible hypotheses for each individual cell must equal one by definition, akin to probabilities.

Prior to receiving any measurements, we have complete uncertainty in our grid occupancy. This uncertainty is captured by setting the mass of set $\left\{F,O\right\}$ to one. Assuming a known mass value for each LiDAR measurement, we can then use Dempster-Shafer's update rule as defined in \cref{eq:dst_rule} to fuse the current belief with the new measurements as they are received.
We update the mass in cell $c$ at time step $k$ by combining the current mass in the cell from previous sensor measurements, $m_{k-1}^{c}$, with the measurement from this time step, $m_{k,z}^{c}$, as follows: 
\begin{align} \label{eq:dst_rule}
 	m_{k}^{c}\left(A\right) &= m_{k-1}^{c} \oplus m_{k,z}^{c}\left(A\right) \\ &\coloneqq \frac{\sum_{X \cap Y = A}m_{k-1}^{c}\left(X\right)m_{k,z}^{c}\left(Y\right)}{1 - \sum_{X \cap Y = \emptyset} m_{k-1}^{c}\left(X\right)m_{k,z}^{c}\left(Y\right)} \nonumber \\ &\qquad \forall A, X, Y \in \left\{\left\{F\right\}, \left\{O\right\}, \left\{F,O\right\}\right\}. \nonumber 
\end{align}
To account for information aging, we employ a discount factor, $\alpha$, to the prior mass in the cell:
\begin{align}
	m_{k,\alpha}^{c}\left(\left\{O\right\}\right) &= \min \left(\alpha \cdot m_{k}^{c}\left(\left\{O\right\}\right), 1\right) \label{eq:m_occ}\\
	m_{k,\alpha}^{c}\left(\left\{F\right\}\right) &= \min \left(\alpha \cdot m_{k}^{c}\left(\left\{F\right\}\right), 1\right) \label{eq:m_free}\\
    m_{k,\alpha}^{c}\left(\left\{F,O\right\}\right) &= 1 - m_{k,\alpha}^{c}\left(\left\{O\right\}\right) - m_{k,\alpha}^{c}\left(\left\{F\right\}\right).
\end{align}
The new masses can then be converted to traditional probabilities using the concept of pignistic probability as follows:
\begin{equation} \label{eq:pignistic}
	betP\left(B\right) = \sum_{A \in 2^{\Omega}} m\left(A\right) \cdot \frac{\vert B \cap A \vert}{\vert A \vert}
\end{equation}
where $B$ is a singleton hypothesis and $\vert A \vert$ is the cardinality of set $A$ (a possible hypothesis) \cite{nuss_particles}.

\subsubsection{Velocity Estimation}
To incorporate dynamics, we estimate the velocity for each cell.
The velocity estimates use a DST approximation for a probability hypothesis density filter with multi-instance Bernoulli (PHD/MIB) \cite{nuss_particles}. DST allows for the particle filter to run more efficiently by initializing particles only in cells with occupied masses above a specified threshold, avoiding occluded regions without measurements. 

Holistically, the PHD/MIB filter initializes particles in the grid by randomly selecting grid cell locations for position and sampling from a zero-mean normal distribution for velocity. Dempster-Shafer's rule then updates the occupancy grid given the propagated particles and the sensor measurement at that time step. An occupancy mass in a cell is equal to the sum of the weights of the particles in the cell. 
The occupancy update is followed by the normalization of the particle weights. A fixed number of new particles is then initialized or ``born" to ensure exploration of the grid space. The mean velocities and occupancies in the grid cells are then stacked into a DOGMa tensor. The particles are resampled from the set of persistent and newly initialized particles for the next particle filter run, maintaining a constant total number of particles across iterations.

\begin{figure}[t!]
	\centering
    \begin{subfigure}{0.6\columnwidth}
		\centering
		\includegraphics[width=0.95\columnwidth, trim = 200 0 200 0, clip]{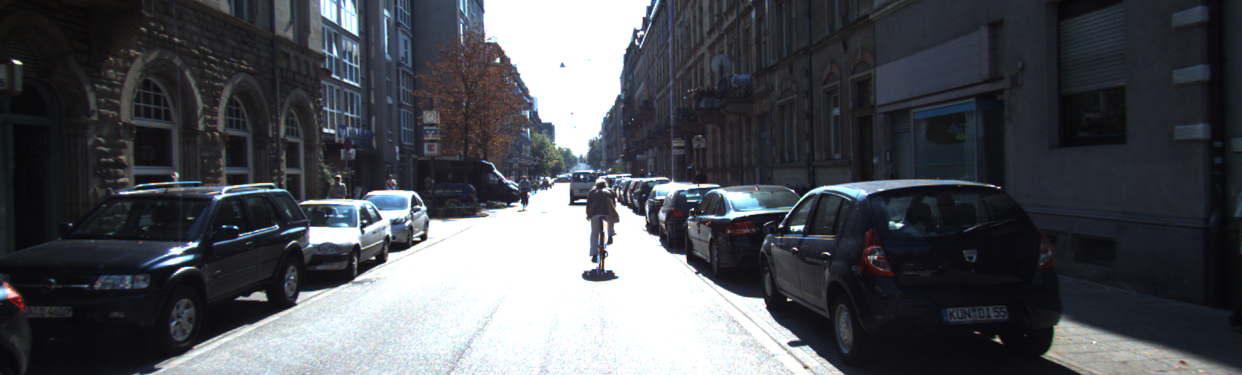}
		\label{fig:camera}
	\end{subfigure}
	\begin{subfigure}{0.35\columnwidth}
		\centering
		\includegraphics[width=0.99\columnwidth, angle = -90, trim = 100 70 170 75, clip]{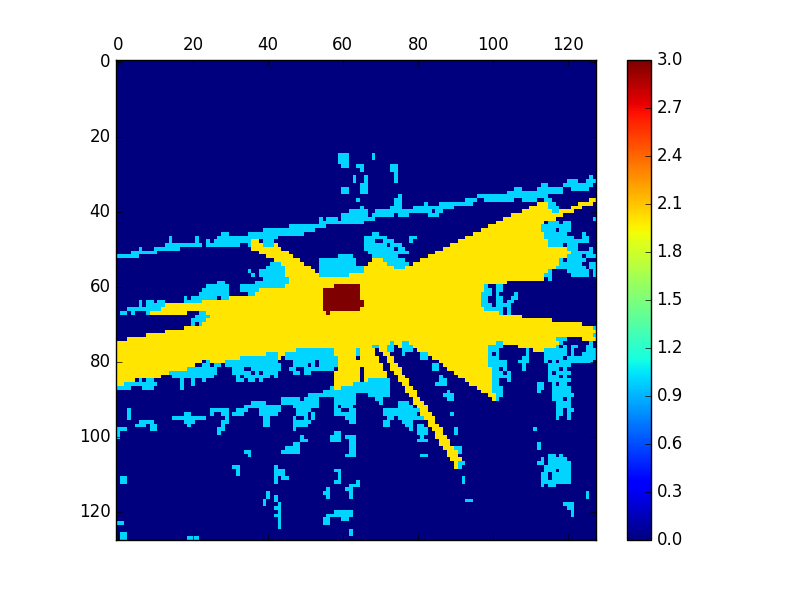}
		\label{fig:example_occ_grid}
	\end{subfigure}

\caption{\small Example scene from the KITTI dataset (left) and corresponding sensor measurement grid (right) showing free (yellow), occupied (light blue), and unknown (dark blue) space \cite{kitti}.} \label{fig:example_data}
\vspace{-10pt}
\end{figure}

\subsection{Neural Network Architecture}
As previously discussed, we pose the environment prediction problem as a video frame prediction task. 
We re-purpose the PredNet architecture to learn the spatial and temporal representation of the environment by training it on occupancy grids instead of images.
The convolutional layers exploit contextual information to correlate the occupied cells, removing the cell independence assumption~\cite{dietmayer}. 
The self-supervised nature of sequential data prediction is advantageous as human-labeled LiDAR data is expensive to obtain~\cite{dietmayer}. 
In this framework, the labels are simply the input environment representation (grids) at a later time instance. Although the original PredNet model was designed for video data, we demonstrate that the architecture can be re-used in the LiDAR setting. The PredNet architecture consists of representation, prediction, target, and absolute error modules. The recurrent representation layer receives absolute error information between the target and the prediction as well as the representation layer output from the next layer. Thus, it updates both laterally and vertically to learn the spatial and temporal internal representation of the data. PredNet employs a regression $l_{1}$-loss for training.

%% file: Sections/04-Experiments.tex
\section{EXPERIMENTS} \label{sec:experiments}

This section describes our dataset generation, implementation, experimental protocol, and baselines used to validate our approach on the KITTI dataset~\cite{kitti}.\footnote{The code implementation of our approach can be found here:\\ \url{https://github.com/mitkina/EnvironmentPrediction}}

\subsection{Dataset Generation}
As shown in \cref{fig:methodology}, the raw sensor data must be transformed into occupancy grids, which are the inputs to our neural network. The following subsections provide details about the data and occupancy grid generation. 

\subsubsection{LiDAR Measurement Grids} \label{sec:data}
The KITTI HDL-64E Velodyne LiDAR dataset was augmented for use in occupancy grid prediction \cite{kitti}. The dataset contains a variety of urban road scenes in Karlsruhe, Germany. We use $35,417$ frames (138 driving sequences) for training, $496$ frames (3 driving sequences) for validation, and $2,024$ frames (7 driving sequences) for testing. Velodyne LiDAR point cloud measurements are obtained at 10 Hz.

Each LiDAR point cloud is filtered to remove the points corresponding to the ground as described in \cref{sec:approach}. Then, a simple form of ray-tracing is performed to determine the free space between a LiDAR measurement and the ego vehicle. Each resulting local grid is centered at the ego vehicle GPS coordinate position. An example of a local grid is shown in \cref{fig:example_data}, where a scene from the dataset is represented as a camera image (left) and as a local occupancy grid (right). The generated grids have a side length of \SI{42.7}{\metre} with \SI{0.33}{\metre} resolution, forming $128 \times 128$ grids. The shorter grid range is acceptable for slower speeds in urban settings, as is the case in the KITTI dataset~\cite{kitti}.

\subsubsection{Dynamic Occupancy Grid Maps} \label{subsubsec:dogma}
The DOGMa's occupancy and velocity information is computed from the LiDAR data as outlined in \cref{sec:approach}. The velocities are then filtered to remove static measurements according to the cell-wise Mahalanobis distance: $\tau = v^{T}Pv,$
where $v$ is the velocity vector and $P$ is the cell's covariance matrix as computed from the particles~\cite{nuss_particles}. Cells with occupancy masses below a threshold are also removed. The velocities are then normalized to the range $\left[-1,1\right]$ and stacked with either (1) the pignistic probability (\cref{eq:pignistic}) or (2) the DST mass (\cref{eq:m_occ}~and~\cref{eq:m_free}) occupancy grids, forming the input to the network. 
\begin{figure}[t!]
	\centering
    \begin{subfigure}{0.49\columnwidth}
		\centering
		\includegraphics[scale=0.29, trim = 3cm 1.5cm 3cm 1.5cm, clip]{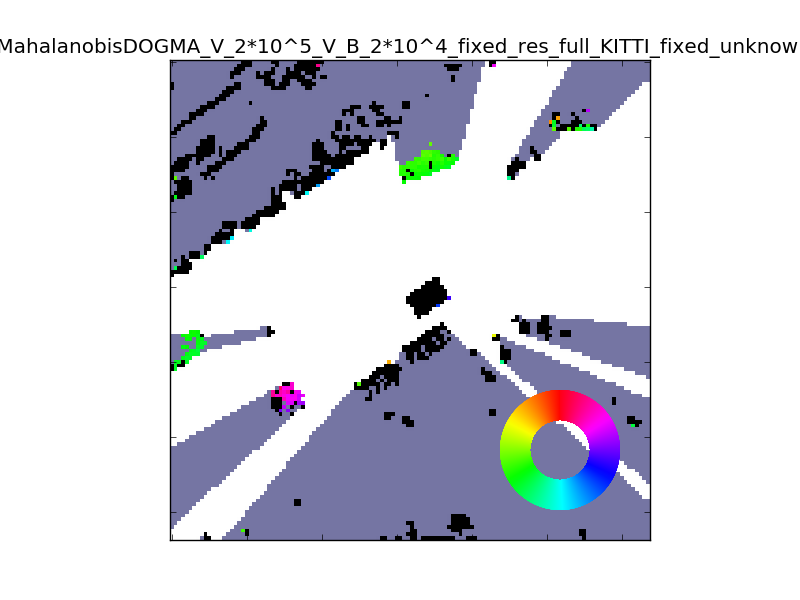}
        \caption{} \label{fig:results_a}
	\end{subfigure}
	\begin{subfigure}{0.49\columnwidth}
		\centering
		\includegraphics[scale=0.29, trim = 3cm 1.5cm 3cm 1.5cm, clip]{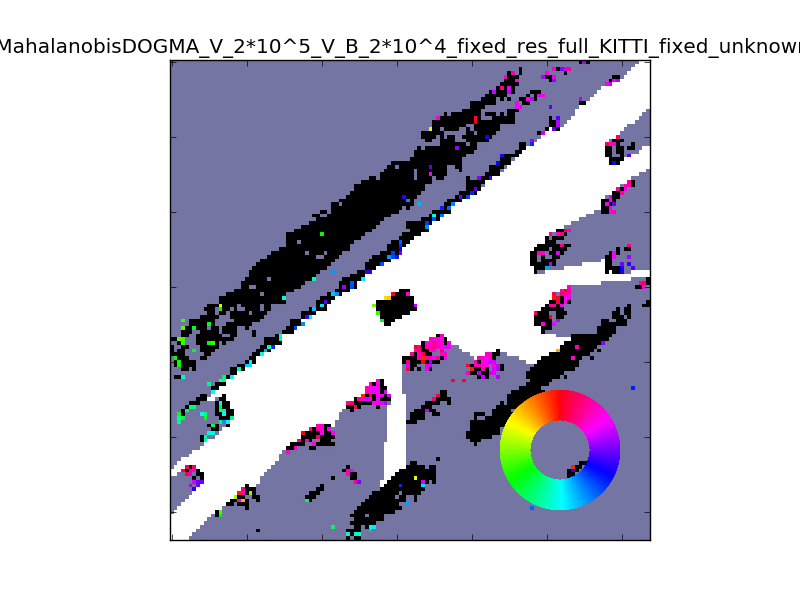}
        \caption{} \label{fig:results_b}
	\end{subfigure}
\caption{\small Examples of DOGMa output on KITTI LiDAR data. Grid cell color represents a local velocity direction corresponding to the color's location on the color wheel.} \label{fig:filter_results}
\vspace{-15pt}
\end{figure}

Example DOGMa velocity data are shown in \cref{fig:filter_results}, with LiDAR static measurement grids as in \cref{fig:example_data} overlaid with velocity information. The ego vehicle is centered in each grid. The colored cells in the grids denote the local velocity direction of those cells. For instance, in \cref{fig:results_a}, the ego vehicle has come to a stop. There is a vehicle traveling in the northeast direction (denoted in purple) behind the ego vehicle, and another two vehicles traveling southwest in the opposing lane (denoted in green). In \cref{fig:results_b}, the ego vehicle is traveling in the southwest direction within the grid. The vehicles denoted in pink are in heavy traffic making them slower than the ego vehicle, and thus appearing to travel in the northeast direction relative to the ego vehicle's frame of reference. The green vehicle is traveling faster than the ego vehicle in the same direction, therefore it has a relative speed in the southwest direction. We note that due to limited computing power, which constrained the number of particles to $2 \times 10^{5}$, the velocities computed by the particle filter are noisy and do not fully capture the dynamic environment surrounding the ego vehicle.

We posit that local as opposed to global velocities in the grids are expected to be sufficient for training of the network since camera data provides local perception information, as well. 
In this manner, all the training, validation, and test data was processed into DOGMas of channel depth three or four respectively (occupancy probability or occupied and free masses and horizontal and vertical velocities in the local grid frame) that serve as input into the PredNet architecture. Since the network is being used to predict into the future, no labels are required other than the existing grids at appropriate future time steps resulting in a self-supervised approach. 

\subsection{PredNet Experiments}
PredNet was trained and tested on an NVIDIA GeForce GTX 1070 GPU. At test time, one sequence (15 predictions totaling \SI{1.5}{\second} ahead) took on average \SI{0.1}{\second} to run.

We train PredNet separately on just the static occupancy grid information versus the full DOGMa data to determine the added benefit (if any) of the supplemental velocity information to the prediction. We also consider the comparison of pignistic probability and DST-based grids. For each set of input data, we tune the network's hyperparameters.
The training protocol considers sequences of 20 grids at a time (at~\SI{10}{\hertz}) and $200$ epochs total with 500 samples per epoch. As suggested by Lotter et al., we train the predictive network in two stages~\cite{prednet}. We first train in the $t + 1$ mode, and then fine-tune in the $t + 5$ mode. In the $t+1$ mode, the network predicts the occupancy grid at the next time frame (\SI{0.1}{\second} ahead). In the $t+5$ mode, the network stops receiving target occupancy grids at time step $5$ (\SI{0.5}{\second}) in the sequence. At this point, it uses its built-up internal representation to recursively predict forward in time (\SI{1.5}{\second} ahead).

\subsection{Baseline Approaches} \label{sec:baselines}

We baseline our results against three other approaches: a static environment assumption, a fully convolutional neural network approach, and particle filter propagation.

As proposed by Lotter et al., we use the last-seen frame baseline, which assumes a static environment. The baseline uses the last target label grid seen by the network as the prediction for future time steps~\cite{prednet}. This baseline follows the naive assumption that the \SI{1.5}{\second} prediction horizon is short enough that the majority of the environment remains static.

Second, we compare our approach to a fully convolutional network that learns spatial features, but does not account for the temporality of the data within its architecture~\cite{schreiber}. This approach is similar to that taken by Hoermann et al.~\cite{dietmayer}. They perform a prediction that also classifies cells as static or moving, while we are simply interested in the cell occupancy. We employ the FCN model~\cite{fcn}, which has shown promising results with DOGMa data~\cite{fcn_baseline}.
We train the predictive model on batches of $50$ probabilistic DOGMas with $1975$ iterations.  
To ensure a fair comparison with PredNet, we employ an $l_{1}$-loss, converting the FCN from a classification to a regression model. To guarantee valid probabilities, we truncate the output weights to between $\left[0,1\right]$.

Lastly, we compare our results to the propagated particle filter prediction. The prediction is formed by propagating the particles in cells above the Mahalanobis distance threshold using a linear dynamics model, and computing the resulting occupancy grid as described in \cref{subsubsec:dogma}. This baseline was also used by Hoermann et al.~\cite{dietmayer}. The assumptions made by this approach are: grid cell independence, linear, noisy particle dynamics, and absence of free space estimation (particles only capture occupied space)~\cite{dietmayer, schreiber}. Due to the last assumption, we use the predicted occupied mass values as a proxy for the occupied probabilities.

%% file: Sections/05-Results.tex
\section{RESULTS} \label{sec:results}

This section summarizes the findings and insights gained from experiments on the KITTI dataset.

\subsection{Quantitative Performance} \label{subsec:cumulative}
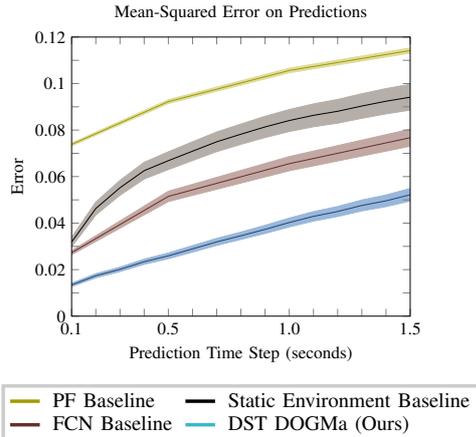
\begin{figure}[t!]
	\centering
	\scalebox{0.65}{\input{new_figs/error_analysis.tex}}
\caption{\small Mean-squared error comparison for all tested approaches in predicting \SI{1.5}{\second} into the future.
} \label{fig:error_plot}
\vspace{-15pt}
\end{figure}

Mean-squared error (MSE) between the predicted and target occupancy grids is used to measure the degree of success for the proposed algorithm. \cref{fig:error_plot} shows the MSE for predictions of up to \SI{1.5}{\second} ahead. The plot demonstrates the PredNet method with DST-based DOGMa inputs, alongside particle filter, static environment assumption, and FCN baselines. For all methods, as the prediction time increases and uncertainty accumulates, the MSE increases. For PredNet, the longer time prediction causes the model to extrapolate the learned environment representation further into the future. Nevertheless, the PredNet model outperforms all other baselines for the \SI{1.5}{\second} time horizon in learning an effective environment motion representation.
As shown in \cref{fig:error_plot}, the PredNet model outperforms the static environment assumption baseline by more than 2 times at \SI{1.5}{\second} predictions, confirming that PredNet successfully learned at least a portion of the dynamics of the environment surrounding the ego vehicle.

We also compare PredNet results to that of an FCN trained on probabilistic DOGMa data, as shown in \cref{fig:error_plot}, to investigate the added benefit of an architecture that incorporates data temporality. PredNet outperforms the FCN; but we note that the FCN surpasses the static environment baseline, showing some ability to learn the environment dynamics.

Lastly, we investigate the performance of the particle filter alone in estimating the future environment state, which results in the poorest model performance. The particle filter is hindered by noisy updates to a relatively low number of particles due to computing resource limitations. As can be seen in \cref{fig:filter_results}, the particle filter is able to distinguish only parts of the environment as moving, causing difficulties in predicting the complete environment propagation in time.

Additionally, we examined the added benefit of DST-based over probability-based occupancy grids as well as DOGMa over static occupancy grid input to the PredNet model. \cref{tab:error} provides the MSE metric across the test dataset for PredNet trained on the four variants of the data. 

As might be expected, PredNet trained on DOGMas slightly outperforms that trained on only occupancy information, especially at longer prediction times. We note that although the network trained on DOGMas is incrementally better than that trained on static grids, both models are within standard error of each other. The DST-based DOGMa outperforms in most prediction time horizons, indicating incremental improvement due to both the DST representation and the velocity information. The subtlety of the added degree of freedom in differentiating lack of information from conflicting information in DST-based grids appears to marginally improve the network prediction capability. However, there seems to be minimal change in performance among all four of the data representations. This result suggests that PredNet is able to learn an effective environment motion representation that performs comparably without the added computational effort of estimating cell velocities, albeit noisy ones.

\begin{table}[b]
\captionsetup{singlelinecheck = false, justification=justified}
\captionof{table}{\small
MSE prediction results at $T$ time steps in the future on the PredNet model trained on static occupancy grids and DOGMas with both probabilistic and DST-based occupancy representations. Bold numbers denote the top performing variant.
}
\normalsize
\label{tab:error}
\begin{center} 
\begin{tabular}{@{}lcccc@{}}
\toprule
 & DST & DST  & Prob. & Prob.  \\
 & Static & DOGMa & Static & DOGMa \\
$T$ & $\times 10^{-3}$ & $\times 10^{-3}$ & $\times 10^{-3}$ & $\times 10^{-3}$ \\
\midrule
\SI{0.1}{\second} & $13.5 \pm 0.7$ & $\mathbf{13.4 \pm 0.7}$ & $\mathbf{13.4 \pm 0.8}$ &  $13.5 \pm 0.7$ \\

\SI{0.5}{\second} & $26.4 \pm 1.3$ & $\mathbf{25.9 \pm  1.2}$ & $26.7 \pm 1.3$ & $26.4 \pm  1.3$ \\

\SI{1.0}{\second} & $40.7 \pm  2.0$ & $40.3 \pm 1.9$ & $40.5 \pm 1.9$ & $\mathbf{40.2 \pm 1.9}$ \\

\SI{1.5}{\second} & $52.6 \pm 2.7$ & $\mathbf{52.2 \pm 2.7}$ & $52.8 \pm 2.7$ & $52.5 \pm 2.7$ \\
\bottomrule
\end{tabular}
\end{center}
\end{table}

\subsection{Qualitative Performance} \label{subsec:visual}
\begin{figure*}[ht]
	\centering
    \includegraphics[scale=0.185]{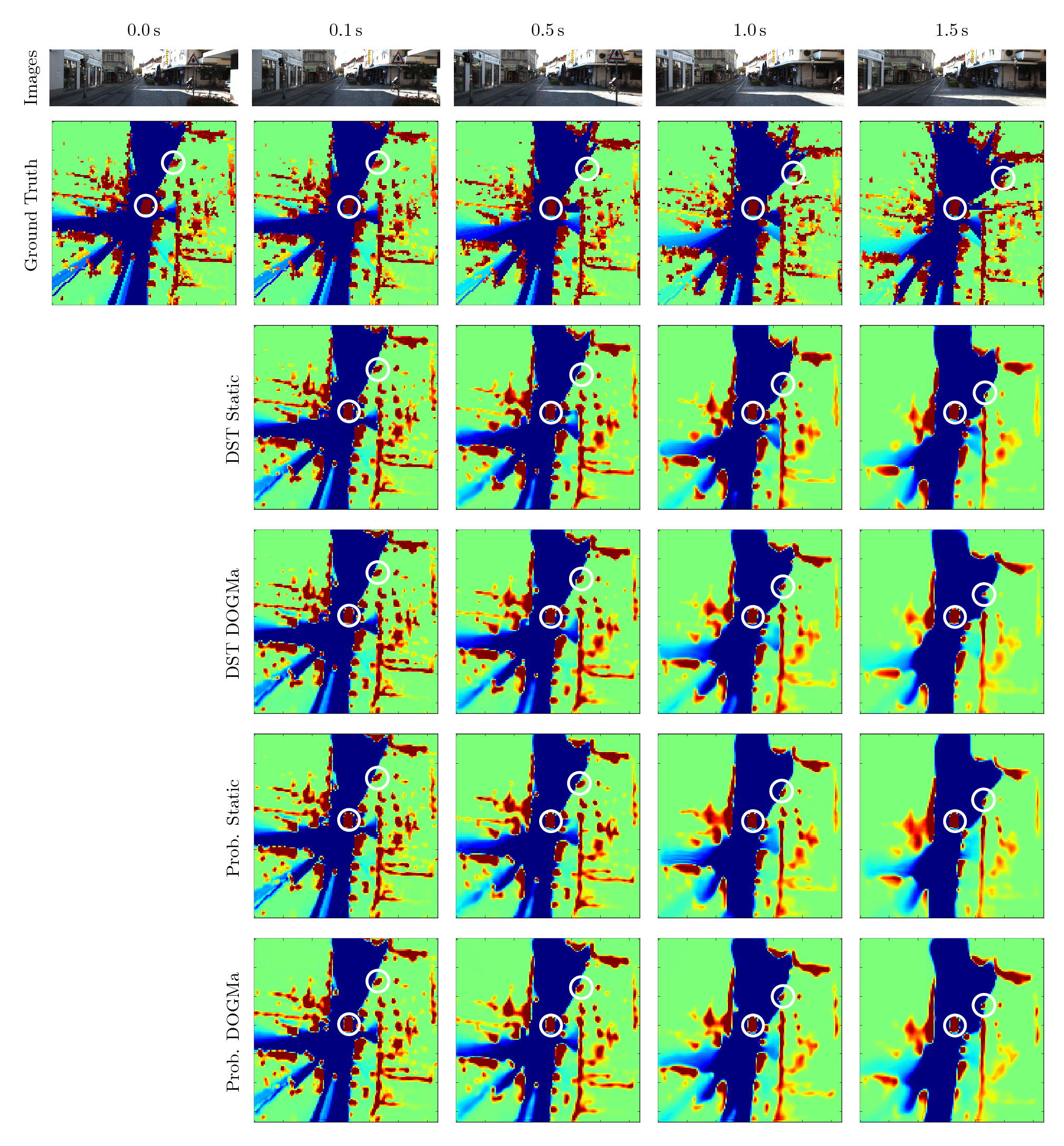}
\caption{\small Overview of occupancy grid predictions in the test set up to \SI{1.5}{\second}. Top row shows camera images of the scene ahead of the ego vehicle. Second row shows the ground truth occupancy grid view of the scene (free: blue, occupied: red, and unknown: green space). The ego vehicle is moving up in the grid (static environment is moving down in the local frame), and a bicyclist ahead of the vehicle is making a right turn (ego vehicle and bicyclist are denoted by white circles). Rows 3--6 show the PredNet predictions on four variants of the input data. The PredNet model was successfully able to capture the dynamic motion of the environment relative to the ego vehicle. At shorter term predictions, the model also captures the turning behavior of the bicyclist.
} \label{fig:results}
\vspace{-15pt}
\end{figure*}
\cref{fig:results} shows an example of the test set output predictions by the PredNet model. The columns denote the prediction time step. The first row shows the camera image depicting the scene ahead of the ego vehicle. The second row shows the corresponding target occupancy grids with the ego vehicle in the center traveling upwards in the grid and a bicyclist ahead of the ego vehicle making a right turn (ego vehicle and bicyclist are denoted by white circles). Rows 3--6 show the predictions made by PredNet when trained on the four variants of the data input. All four input data variants were successfully able to predict the relative motion of the static environment with respect to the ego vehicle (in the downward direction of the grid). 

Visually, the probabilistic grids (particularly the static occupancy ones) show lower quality predictions at \SI{1.5}{\second}, resulting in blurring and disappearance of objects from the environment. The DST-based grids were able to maintain more of the structure of the environment at longer prediction horizons. However, the difference between the static and DOGMa data is less distinct. It visually appears that the DST-based static grid outperforms the DST-based DOGMa. The static grid PredNet mainted a slightly sharper result, and was able to predict the turning motion of the bicyclist to a higher degree of success than the DOGMa PredNet. Nevertheless, both DST outputs resulted in comparable predictions. We note that PredNet appears to blur the predictions further away from the ego vehicle (center of grid), which is in agreement with the LiDAR measurement uncertainty increasing further away from the sensor.

\subsection{Discussion}
We present these results with the caveat that the particle filter methodology is only able to provide noisy estimates of the cell velocities, and particle filter performance is limited by available computational power. Concurrent work by Schreiber et al. shows some improvement with the addition of velocity information, but the work does not provide error bounds or statistical analysis on the significance of the findings~\cite{schreiber}.
Although PredNet successfully predicted the motion of the static environment relative to the ego vehicle, the model was only able to make accurate predictions of the obstacles in the environment over a shorter time horizon. In this work, we aimed to push the limits of a model-free video frame prediction style approach. For more robust predictions, it may be useful to use object detection algorithms to enrich the input data. The separation of static and dynamic data within the neural network architecture~\cite{schreiber} may also decrease the detrimental blurring and disappearance of obstacles at higher prediction horizons. Furthermore, a deterministic model such as PredNet is inherently limited in its prediction capability due to the underlying multimodal distribution over the potential futures. Recent advances in generative modeling for prediction provide a means to address the multimodality without averaging the potential predictions, suggesting a promising avenue for future work~\cite{rui_cvae}.

%% file: new_figs/error_analysis.tex
\begin{tikzpicture}

\definecolor{color1}{rgb}{0.917647058823529,0.607843137254902,0.603921568627451}
\definecolor{color0}{rgb}{0.713725490196078,0.686274509803922,0.662745098039216}
\definecolor{color3}{rgb}{0.686274509803922,0.886274509803922,0.92156862745098}
\definecolor{color2}{rgb}{0.725490196078431,0.835294117647059,0.725490196078431}
\definecolor{color5}{rgb}{0.627450980392157,0.141176470588235,0.133333333333333}
\definecolor{color4}{rgb}{0.6,0.717647058823529,0.858823529411765}
\definecolor{color7}{rgb}{0.223529411764706,0.717647058823529,0.803921568627451}
\definecolor{color6}{rgb}{0.329411764705882,0.545098039215686,0.329411764705882}
\definecolor{color8}{rgb}{0.149019607843137,0.274509803921569,0.427450980392157}
\definecolor{color9}{rgb}{0.760,0.658,0.639}
\definecolor{color10}{rgb}{0.396,0.184,0.184}
\definecolor{color11}{rgb}{0.627, 0.607, 0.011}
\definecolor{color12}{rgb}{0.874, 0.862, 0.607}

\definecolor{color1}{rgb}{0.686274509803922,0.886274509803922,0.92156862745098}
\definecolor{color0}{rgb}{0.713725490196078,0.686274509803922,0.662745098039216}
\definecolor{color3}{rgb}{0.223529411764706,0.717647058823529,0.803921568627451}
\definecolor{color2}{rgb}{0.6,0.717647058823529,0.858823529411765}
\definecolor{color4}{rgb}{0.149019607843137,0.274509803921569,0.427450980392157}
\definecolor{color5}{rgb}{0.760,0.658,0.639}
\definecolor{color6}{rgb}{0.396,0.184,0.184}
\definecolor{color11}{rgb}{0.627, 0.607, 0.011}
\definecolor{color12}{rgb}{0.874, 0.862, 0.607}

\begin{axis}[
title={Mean-Squared Error on Predictions},
name=plot2,
anchor=south,
xlabel={Prediction Time Step (seconds)},
ylabel={Error},
xmin=1, xmax=15,
ymin=0.0, ymax=0.12,
axis on top,
ytick={0.0,0.01,0.02,0.03,0.04,0.05,0.06,0.07,0.08,0.09,0.1,0.11, 0.12},
yticklabels={0,,0.02,,0.04,,0.06,,0.08,,0.1,,0.12},
scaled y ticks = false,
xtick={1,2,3,4,5,6,7,8,9,10,11,12,13,14,15},
xticklabels={0.1,,,,0.5,,,,,1.0,,,,,1.5},
tick pos=both,
legend cell align={left},
legend entries={{PF Baseline},{Static Environment Baseline},{FCN Baseline},{DST DOGMa (Ours)}},
legend columns=2,
legend style={font=\fontsize{12}{5}\selectfont, column sep=1ex, at={((0.5,-0.45)},anchor=south, draw=white!80.0!black, line width=2pt}
]
\addlegendimage{no markers, color11}
\addlegendimage{no markers, black}
\addlegendimage{no markers, color6}
\addlegendimage{no markers, color3}
\addlegendimage{no markers, color4}

\path [draw=color0, fill=color0] (axis cs:1,0.0338986702263355)
--(axis cs:1,0.0298364721238613)
--(axis cs:2,0.0435115769505501)
--(axis cs:3,0.0519944429397583)
--(axis cs:4,0.0589604452252388)
--(axis cs:5,0.0630328729748726)
--(axis cs:6,0.0668914541602135)
--(axis cs:7,0.0707276538014412)
--(axis cs:8,0.0738609954714775)
--(axis cs:9,0.0767842531204224)
--(axis cs:10,0.0793823376297951)
--(axis cs:11,0.0814220979809761)
--(axis cs:12,0.0829752534627914)
--(axis cs:13,0.0850551500916481)
--(axis cs:14,0.0869568362832069)
--(axis cs:15,0.0885869562625885)
--(axis cs:15,0.0996733903884888)
--(axis cs:15,0.0996733903884888)
--(axis cs:14,0.0978498831391335)
--(axis cs:13,0.0956215634942055)
--(axis cs:12,0.0931855440139771)
--(axis cs:11,0.091255895793438)
--(axis cs:10,0.088833399116993)
--(axis cs:9,0.0859259516000748)
--(axis cs:8,0.0825875625014305)
--(axis cs:7,0.0791909471154213)
--(axis cs:6,0.0749591365456581)
--(axis cs:5,0.0706620737910271)
--(axis cs:4,0.0661514922976494)
--(axis cs:3,0.058403342962265)
--(axis cs:2,0.0490802153944969)
--(axis cs:1,0.0338986702263355)
--cycle;

\path [draw=color2, fill=color2] (axis cs:1,0.0141664780676365)
--(axis cs:1,0.0126944575458765)
--(axis cs:2,0.0164950545877218)
--(axis cs:3,0.0191677939146757)
--(axis cs:4,0.0222989898175001)
--(axis cs:5,0.0246997252106667)
--(axis cs:6,0.0275362487882376)
--(axis cs:7,0.0304189249873161)
--(axis cs:8,0.0329827964305878)
--(axis cs:9,0.0355606004595757)
--(axis cs:10,0.0383509732782841)
--(axis cs:11,0.0408631712198257)
--(axis cs:12,0.042839378118515)
--(axis cs:13,0.0452423244714737)
--(axis cs:14,0.0470754504203796)
--(axis cs:15,0.0494950413703918)
--(axis cs:15,0.0549006387591362)
--(axis cs:15,0.0549006387591362)
--(axis cs:14,0.0520861819386482)
--(axis cs:13,0.0499449521303177)
--(axis cs:12,0.0471439883112907)
--(axis cs:11,0.0449425727128983)
--(axis cs:10,0.0421728082001209)
--(axis cs:9,0.0390626341104507)
--(axis cs:8,0.0361415073275566)
--(axis cs:7,0.0334457829594612)
--(axis cs:6,0.030316973105073)
--(axis cs:5,0.027172576636076)
--(axis cs:4,0.0245015751570463)
--(axis cs:3,0.0210831891745329)
--(axis cs:2,0.0182703752070665)
--(axis cs:1,0.0141664780676365)
--cycle;

\path [draw=color5, fill=color5] (axis cs:1,0.028171377198305)
--(axis cs:1,0.0264138151542284)
--(axis cs:5,0.0492665790952742)
--(axis cs:10,0.062500115018338)
--(axis cs:15,0.0730411987751722)
--(axis cs:15,0.0803894121199846)
--(axis cs:15,0.0803894121199846)
--(axis cs:10,0.0686057996936142)
--(axis cs:5,0.0537026501260698)
--(axis cs:1,0.028171377198305)
--cycle;

\path [draw=color12, fill=color12] (axis cs:1,0.0744546038568512)
--(axis cs:1,0.0732097940980896)
--(axis cs:5,0.0914011235429581)
--(axis cs:10,0.104625612310633)
--(axis cs:15,0.113082021829451)
--(axis cs:15,0.115304082277452)
--(axis cs:15,0.115304082277452)
--(axis cs:10,0.106615841575399)
--(axis cs:5,0.0930652980611984)
--(axis cs:1,0.0744546038568512)
--cycle;

\addplot [black]
table {%
1 0.0318675711750984
2 0.0462958961725235
3 0.0551988929510117
4 0.0625559687614441
5 0.0668474733829498
6 0.0709252953529358
7 0.0749593004584312
8 0.078224278986454
9 0.0813551023602486
10 0.084107868373394
11 0.086338996887207
12 0.0880803987383842
13 0.0903383567929268
14 0.0924033597111702
15 0.0941301733255386
};

\addplot [color4]
table {%
1 0.0134304678067565
2 0.0173827148973942
3 0.0201254915446043
4 0.0234002824872732
5 0.0259361509233713
6 0.0289266109466553
7 0.0319323539733887
8 0.0345621518790722
9 0.0373116172850132
10 0.0402618907392025
11 0.042902871966362
12 0.0449916832149029
13 0.0475936383008957
14 0.0495808161795139
15 0.052197840064764
};

\addplot [color6]
table {%
1 0.0272925961762667
5 0.051484614610672
10 0.0655529573559761
15 0.0767153054475784
};

\addplot [color11]
table {%
1 0.0738321989774704
5 0.0922332108020782
10 0.105620726943016
15 0.114193052053452
};

\end{axis}


\end{tikzpicture}

%% file: Sections/06-Conclusion.tex
\section{CONCLUSIONS} \label{sec:conclusions}

We present an occupancy grid prediction framework for cluttered, urban environments with dynamic obstacles. We show that by framing environment prediction as a video frame prediction task, we may neglect estimating the environment dynamics within the perception pipeline, without significant loss of accuracy. Experimental results suggest that the Dempster-Shafer evidential grid representation provides slightly more robust training data to the neural network than probabilistic occupancy grids. 
The ConvLSTM successfully learns an effective internal representation of the environment dynamics, outperforming fully convolutional approaches. The proposed perception system goes beyond standard sensor measurements by predicting the temporal evolution of the environment, which can then be used by a path planner to generate smoother, more robust trajectories.

While the results of this framework are promising, we hope to expand this work to longer prediction time horizons. In future work, we aim to better capture uncertainty and confidence in the prediction using conditional variational autoencoders as proposed by others~\cite{finn_cvae, freeman_cvae} and to test the resulting perception module in an autonomous vehicle pipeline to quantify the impact on path planning.

%% file: Sections/07-Acknowledgements.tex
\section{ACKNOWLEDGMENT}
\label{sec:acknowledgements}

The authors would like to acknowledge this project being made possible by the funding from the Ford-Stanford Alliance. We thank Michael Anderson and Henry Shi for help implementing the particle filter, and Bill Lotter for the detailed email discussions, which provided valuable insight into the PredNet architecture.